# Simplified Learning of CAD Features Leveraging a Deep Residual Autoencoder

Raoul Schönhof[a,*], Jannes Elstner[a], Radu Manea[a],

Steffen Tauber[b], Ramez Awad[a], Marco F. Huber[a,c]

[a]*Fraunhofer Institute for Manufacturing Engineering and Automation IPA, Nobelstraße 12, Stuttgart, Germany*
[b]*evia solutions GmbH & evia consulting GmbH, Stuttgart, Am Wallgraben 100, Germany*
[c]*Institute of Industrial Manufacturing and Management IFF, University of Stuttgart, Allmandring 35, Stuttgart, Germany*

* Corresponding author. E-mail address: raoul.schoenhof@ipa-extern.fraunhofer.de

**Abstract**

In the domain of computer vision, deep residual neural networks like EfficientNet have set new standards in terms of robustness and accuracy. One key problem underlying the training of deep neural networks is the immanent lack of a sufficient amount of training data. The problem worsens especially if labels cannot be generated automatically, but have to be annotated manually. This challenge occurs for instance if expert knowledge related to 3D parts should be externalized based on example models. One way to reduce the necessary amount of labeled data may be the use of autoencoders, which can be learned in an unsupervised fashion without labeled data. In this work, we present a deep residual 3D autoencoder based on the EfficientNet architecture, intended for transfer learning tasks related to 3D CAD model assessment. For this purpose, we adopted EfficientNet to 3D problems like voxel models derived from a STEP file. Striving to reduce the amount of labeled 3D data required, the networks encoder can be utilized for transfer training.

## 1. Introduction

Increasing production costs is a key challenge in a modern product design pipeline. This accounts especially to the manufacturing costs, since around 70% of these costs are defined during the early phase of product design [1]. Tackling this challenge, numerous methods, e.g. design for automation or automation potential analysis, have been developed and applied successfully. However, these methods still depend on a human expert for assessing given design features of the product and the CAD model, respectively. Even though CAD systems can assist within a variety of different design optimization techniques [2], assessing abstract features like the ability to automatically feed a component in a manufacturing process, is still a challenging task [3]. With the rise of artificial intelligence and especially deep neural networks, it would be beneficial to assist these experts by creating a neural network based classifier, as proposed in our previous work for automation capability assessment of CAD parts [3][4].

Anyhow, when it comes to machine learning, a key factor for developing a performant model is a sufficient amount of training data. This is an exceptional problem because labeled data for exotic problems, e.g. the automation capabilities of a product, do not exist. Labeling enough data to train a model from scratch would quickly become uneconomical, which is why a pretrained autoencoder can be used to reduce the need of labeled data. This process is well understood. As long the domain of the input does not change, the encoder provides a set of initial weight for a variety of different learning task [5].

In the domain of 2D computer vision, several pretrained high performant models are already available, e.g. ResNet [6], Inception [7], NASNet [8] or EfficientNet [9][10]. Also pretrained point cloud based models are available, e.g. PointNet [11]. Unfortunately, the number of models, which can process voxel models is sparse. To overcome this shortcoming, in this paper we adapted the architecture of EfficientNet, being one of the best known architectures for image classification, to a 3D voxel-based autoencoder. Thus, creating an efficiently trainable base model intended for transfer learning tasks related to 3D CAD model assessment.

The paper is structured as follows: Chapter 2 depicts a brief overview about currently available models, especially the EfficientNet Model. Chapter 3 then shows how to adapt the EfficientNet architecture to be trained in an unsupervised fashion on 3D voxel models from a given unlabeled data set. The autoencoder model is than evaluated against a standard deep convolutional network. The paper closes with a brief summary and an outlook to future work.

## 2. State of the Art

In contrast to 2D models, where only Keras already supports around 26 pretrained networks [10], the number of available pretrained 3D models is sparse. The most relevant publicly available architectures are briefly described in the following.

### 2.1. Generative and Discriminative Voxel Modeling with Convolutional Neural Networks

This work represents one of the first 3D voxel-based autoencoders. The system consists of a 3D convolutional neural network (CNN) / 3D deconvolutional neural network (DCNN) architecture. The overall number of layers is 15; six layers were used for the encoder, one layer for the latent space and the remaining six for the decoder. Within the network, the number of filters used for each layer grew gradually towards the latent space. The last layer of the encoder as well as the first layer of the decoder used fully connected layers with 343 neurons. The latent space itself comprised 100 neurons [13]. Additionally to the autoencoders, Brock et al. also proposed 3D variants of the InceptionNet [7] and the ResNet [4] architecture for classification purposes [13]. However, similar to most progressively growing CNN-DCNN network architectures, the training is computationally intensive. The training code of the network without its weights is available for download [14].

### 2.2. Weighted-Voxel

The weighted-voxel model is a more recent architecture. In contrast to Brock et al., weighted-voxel does not depend on 3D CNNs. Instead, it leverages multi-view projections to feed a series of 2D inputs, which are fed through a 2D autoencoder [15]. The projections then are reconstructed as voxel model. In order to preserve information between the single views, the latent space is defined as an LSTM layer. Besides utilizing multi-view projections and an LSTM layer, the 2D encoder and decoder are still comprised by a conventional 2D CNN-DCNN architecture [15]. The training code of the network without its weights is available for download [16].

### 2.3. EfficientNet

EfficientNets [9] are a family of deep convolutional neural networks popular in image classification. The architecture is similar to MnasNet [9] and was developed by neural architecture search, optimizing accuracy and floating point operations per second (FLOPS) of the network.

The main building block of the EfficientNet is the mobile inverted bottleneck MBConv to which squeeze-and-excite optimization was added [9].

The result of the neural architecture search is the baseline EfficientNetB0. To obtain a family of networks, this baseline is scaled up using compound scaling, a novel scaling method the authors proposed. Compound scaling involves scaling depth/width/resolution of a network uniformly instead of independently/arbitrarily.

Compared to other convolutional neural networks such as SENet, NasNet-A or Inception-v2, EfficientNets are smaller, computationally cheaper and faster [8], which makes them more suitable for working with 3D data.

To our knowledge, EfficientNets have only been used for 2D data, but they can easily be adapted to 3D data by replacing the 2D operations with the respective 3D operations.

### 2.4. Datasets

While the number of labeled trainset is quite restricted, there are numerous unlabeled trainsets for unsupervised training applications. The *ABC Trainset* belongs to the most important datasets in the domain of 3D CAD models. With 1.000.000+ parts, the ABC-Dataset is large enough for statically learning deep networks [17]. An advantage of the ABC Dataset is its focus on mechanical parts with sharp edges and defined surfaces [17]. However, the lack of semantic information requires unsupervised training techniques, e.g. autoencoder.

## 3. Proposed Work

### 3.1. Data Preparation (Voxelization)

The data given in the ABC dataset are stored in a STEP format; in order to process them with a neural network, they have to be converted into a discrete signal, thus the models have to be voxelized. This was done as described in our previous work by utilizing a STEP-STL converter followed by the binvox [18][19] converter.

### 3.2. Voxel-based Deep Residual Network

For the encoder, we use a standard EfficientNet [9] with the fully connected layer serving as the bottleneck of the autoencoder.

The decoder is based on the EfficientNet and is an inverted version of the encoder as well. Compared to the encoder, the order of blocks is reversed and the amount of input filters and output filters of each block is switched.

The main building block of the encoder is the mobile inverted bottleneck MBConv [9] as mentioned above. In order to revert the MBconv operation in the decoder, we introduce a block called MBConvTranspose. Compared to MBConv, MBConvTranspose replaces the depthwise convolutions with depthwise transposed convolutions. This is necessary to allow upsampling via strided depthwise transposed convolution. In addition, the expand layer in the MBConvTranpose expands the channels from $C_{in}$ to $4 \cdot C_{out}$ instead of $4 \cdot C_{in}$ as in the MBConv. This balances the model sizes of encoder and decoder.

Other details are unchanged from the original EfficientNet [9], e.g., using Sigmoid Linear Units (SiLU) activations, batch normalization and squeeze-and-excite optimization.

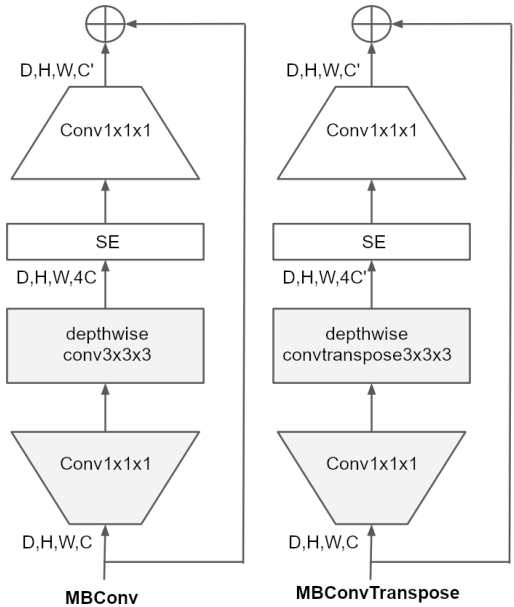

Figure 1: Structure of MBConv3D and MBConvTranspose3D as extension of EfficentnetV2 proposed by Mingxing/Quoc [20]

As we operate on three-dimensional data and require an encoder and decoder, respectively, being much smaller than even the EfficientNetB0, we utilize a custom configuration for the network. Since going from 2D to 3D changes the width, depth and resolution scaling parameters for the EfficientNet, we do not use these scaling parameters to scale the EfficientNetB0 down as proposed in the EfficientNet-eLite approach [21].

## 4. Evaluation

### 4.1. Baseline

For comparison purposes, we use an autoencoder based on conventional convolutional layers. Convolutional neural networks are well known for extracting features of images and converting the images into lower dimensional representations. The purpose of convolutional layers in the autoencoder is to encode the information into a low dimensional representation, which is the feature map of the autoencoder. For the decoder part, we use convolutional transposed layers, which upsample the low dimensional information to the original size. In order to encode and shrink the size of the input, we use a combination of convolution and max pooling layers. All the layers used for the CNN-autoencoder are adapted to work with data in 3D. The architecture of the used CNN encoder is described in Table 1. The convolution blocks in the autoencoder encode different aspects of the 3D objects while the max-pooling layers reduce the size of the data.

By gradually shrinking the size of the data, we are able to encode the information of $64^3$ voxels, therefore approx. $2.62*10^5$ data points into a latent space comprised by 256 data points. The decoder part consists of convolution-transposed layers, which are alternated by dropout layers similar to the encoder. The encoder contains 81K trainable parameters, while the decoder contains 86K trainable parameters. This leads to a reference DCNN autoencoder with 167K parameters, as shown in Table 1.

Table 1: Encoder Architecture

| Rep. | Layer | Output Shape | Kernel |
|---|---|---|---|
| 3 | Conv3D | (64,64,64,4) | 3x3x3 |
| 1 | MaxPool3d | (32,32,32,4) | 2x2x2 |
| 1 | Dropout | (32,32,32,4) | - |
| 2 | Conv3D | (32,32,32,8) | 3x3x3 |
| 1 | MaxPool3d | (16,16,16,8) | 2x2x2 |
| 1 | Conv3D | (16,16,16,8) | 3x3x3 |
| 1 | Conv3D | (16,16,16,16) | 3x3x3 |
| 1 | MaxPool3d | (8,8,8,16) | 2x2x2 |
| 2 | Conv3D | (8,8,8,32) | 3x3x3 |
| 1 | MaxPool3d | (4,4,4,32) | 2x2x2 |
| 5 | Conv3D | (4,4,4,32) | 3x3x3 |
| 1 | Conv3D | (4,4,4,4) | 3x3x3 |
| 1 | Dropout | (4,4,4,6) | - |

The decoder is structured in the opposing order.

### 4.2. Training Setup

Both networks have been trained on the ABC dataset for three epochs. The dataset is split into a trainset comprising 790.176 models and a testset comprising 197.545 models. Both networks were trained with an ADAM optimizer and a learning rate of $1 \cdot 10^{-3}$.

In both cases, the batch size was set to utilize the whole GPU memory available. In our case, we used 11 GB of a single Nvidia GTX 1080 Ti, leading to a batch size of 64 for the reference model as well as our residual autoencoder. As loss function, Mean-Squared-Error was chosen.

### 4.3. Results

In terms of output quality, the reference network was evaluated on 198K models. It achieved a validation score of $1,895 \cdot 10^{-2}$ after six epochs, our network achieved a validation score of 8,780·10-3. Therefore, our residual network was able to achieve a performance increase compared to a reference CNN-DCNN autoencoder. Figure 2 shows the achieved train loss in relation to the training steps.

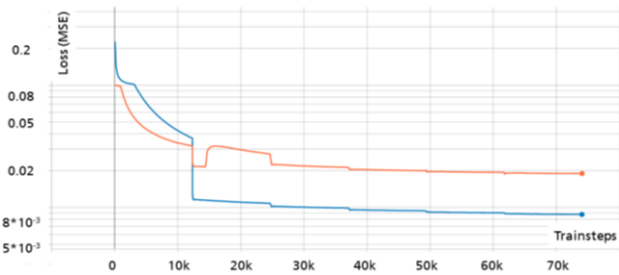

Figure 2: Comparisons of the batch loss (MSE) and the training progress (steps) between the reference DCNN autoencoder (orange) and our residual autoencoder (blue) after 6 epochs (74k steps) of training.

In comparison, the CNN-DCNN autoencoder required approx. 2,93 hours to complete an epoch, while our model required only 1,03 hours. This results in a computational speedup of 284% for a similar number of network weights. The difference in computational efficiency is improved, when comparing the train loss to the computational time, as shown in Figure 3.

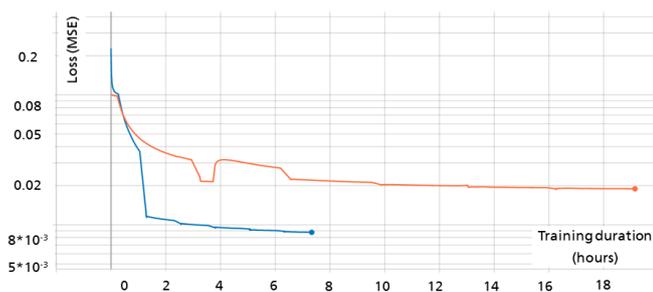

Figure 3: Comparisons of the batch loss (MSE) and the training time (hours) between the reference DCNN autoencoder (orange) and our residual autoencoder (blue) after 6 epochs (74k steps) of training.

For illustration, Input-Output examples of 12 models are given in the appendix. The encoder of the network was then used for transfer learning in the domain of automation assessment as described in our previous work [3]. Here we use the encoder network to assess assembly relevant features, whereas our dataset is extremely small (approx. 220 hand labeled models). Without such a pretraining, a big network as described would not be able to generalize and just overfit.

5. **Conclusion**

Within this work, we presented a deep residual autoencoder network, based on the EfficientNet architecture, for unsupervised training on voxel data. Furthermore, we showed that our network converged much faster compared to a conventional 3D-CNN-DCNN based architecture when trained on the ABC-trainset. We showed that our network achieved a better performance and faster training as a conventional 3D-CNN-DCNN architecture with the same number of weights. This makes our network ideal for transfer learning purposes in conditions where little amount of labeled data is available. The system, including a pretrained network, is publicly available on GitHub [22], providing an encoder suitable for various projects requiring transfer learning.

While our current network is intentionally small, the next step would be to increase the depth and width of the MBConv3D and MBConvTranspose3D blocks. This would allow to preserve even more information within the latent space. As well, the decreased need for computational resources could allow us to increase the resolution of the voxel model in order to catch details, which would otherwise not be able to be represented in a voxel model.

Finally, we aim to analyze the autoencoder results with explainable AI methods in order to further estimate its quality.

**Acknowledgements**

The research presented has received funding from the German Federal Ministry for Education and Research (BMBF/PTKA) within the project "KiMont" (Grant number 02K20K512).

**Appendix: Examples**

Table 2: Input-Output evaluation examples of our EfficentNet-based 3D autoencoder trained on 790K models from the ABC trainset. Latent space: 256 Neurons, resolution: $64^3$ voxels per model.

| Input 1 | Output 1 |
|---|---|
| 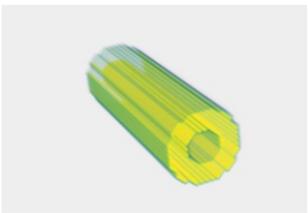 | 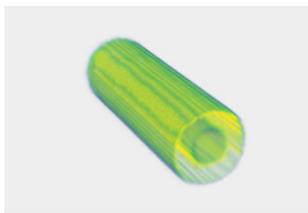 |
| 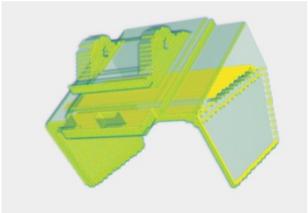 | 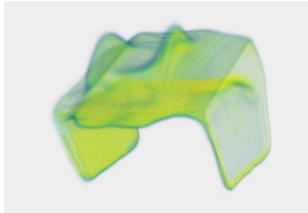 |
| 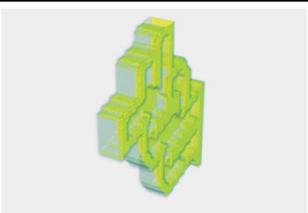 | 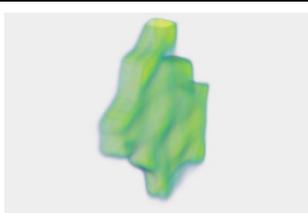 |
| 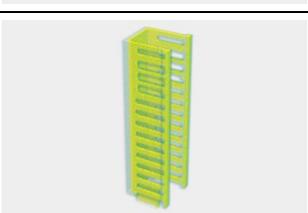 | 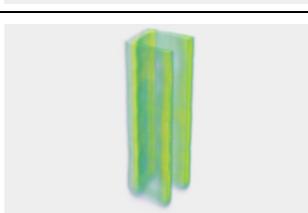 |
| 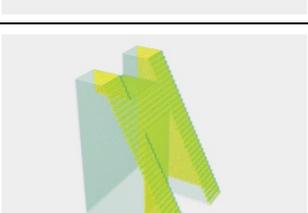 | 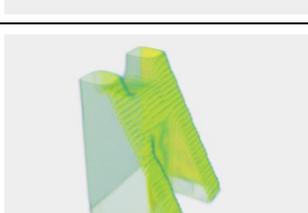 |

| Input 2 | Output 2 |
|---|---|
| 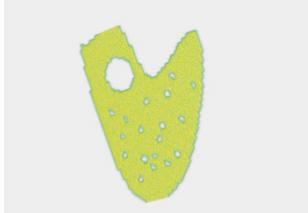 | 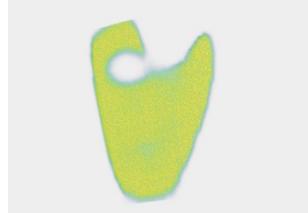 |
| 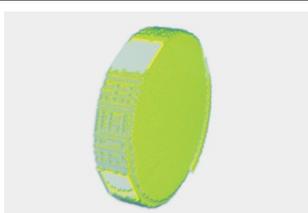 | 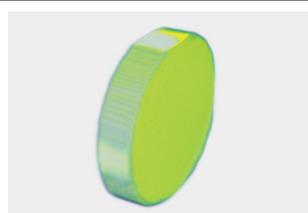 |
| 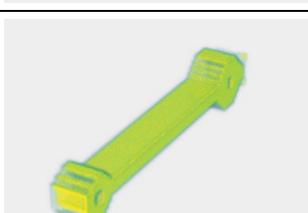 | 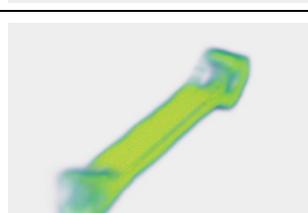 |
| 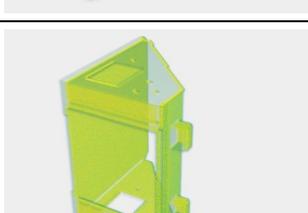 | 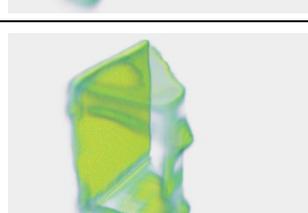 |
| 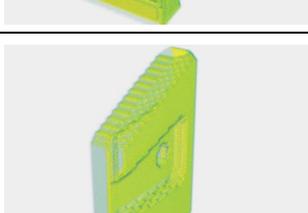 | 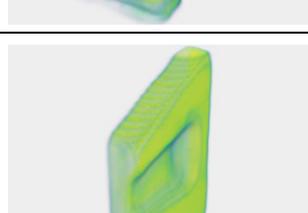 |
| 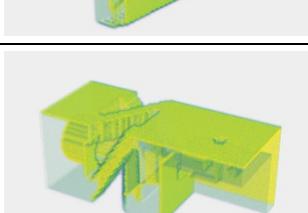 | 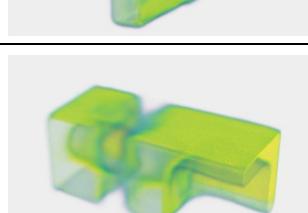 |
| 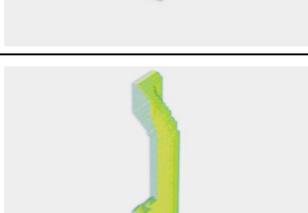 | 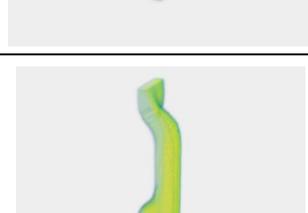 |